%% file: FedKMeansGTVMin_Arxiv.tex
\documentclass[conference]{IEEEtran}
\IEEEoverridecommandlockouts

\usepackage{amsmath,amssymb,amsfonts}
\usepackage{graphicx}
\usepackage{booktabs}
\usepackage{cite}
\usepackage{hyperref}
\usepackage{algorithm}
\usepackage{algpseudocode} 
\usepackage{amsmath}
\usepackage{bm}
\usepackage{amsthm}
\usepackage{subcaption}
\usepackage{ifplatform}

\usepackage{todonotes}

\usepackage{pgffor} 
\usepackage{pgfplots, pgfplotstable}
\usepgfplotslibrary{fillbetween}
\pgfplotsset{compat=newest}
\usepgfplotslibrary{groupplots}  
\pgfplotscreateplotcyclelist{mycolors}{
	{blue, mark=*},
	{red, mark=square*},
	{green!50!black, mark=triangle*},
	{black, mark=o},
}

\usepackage{tikz} 
\usetikzlibrary{positioning,arrows.meta,calc}
\algrenewcommand{\algorithmicensure}{\textbf{Output:}}

\usepackage{siunitx}

\input{ml_macros.tex}

\title{Federated \emph{k}-Means over Networks}

\author{
Xu~Yang\textsuperscript{*}, Salvatore~Rastelli\textsuperscript{*}, and Alexander~Jung\\
Department of Computer Science, Aalto University, Finland\\
E-mails: \{xu.1.yang, salvatore.rastelli, alexander.jung\}@aalto.fi\\
\textsuperscript{*}These authors contributed to this work equally.
}

\begin{document}
	\maketitle	
	
	\begin{abstract}
		We study federated clustering, where inter-connected devices collaboratively 
		cluster the data points of private local datasets. Focusing on hard clustering 
		via the $k$-means principle, we formulate federated $k$-means as an instance of generalized total variation minimization (GTVMin). 
		This leads to a federated $k$-means algorithm in which each device updates 
		its local cluster centroids by solving a regularized $k$-means problem  
		with a regularizer that enforces consistency between neighbouring devices. 
		The resulting algorithm is privacy-friendly, as only aggregated information is exchanged.
	\end{abstract}
	
	\section{Introduction}

	Clustering is one of the most fundamental and widely-used techniques for data analysis and 
	unsupervised machine learning (ML) \cite{MLBasics,hastie01statisticallearning,AaltoDictofML}. 
	The goal of clustering is to decompose a given dataset into coherent sub-groups 
	or clusters \cite{MLBasics,hastie01statisticallearning,BishopBook}. Our 
	focus is on hard clustering where the clusters are non-overlapping, i.e., they 
	form a partition of the dataset. 
	
	A widely used clustering method for Euclidean feature vectors is $k$-means \cite[Sec.~8.1]{MLBasics}. 
	The method partitions data points into $\nrcluster$ clusters by minimizing the 
	average squared Euclidean distance to the nearest cluster centroid. As this problem is 
	NP-hard in general \cite{Aloise:2009aa}, practical solutions rely on (variations of) Lloyd's algorithm  \cite{Lloyd1982,Gray1980}.
	
	Standard implementations of $k$-means methods require centralized access to the 
	full dataset. This can be infeasible when data are generated and stored 
	across multiple devices with private local datasets. Central aggregation 
	may be prohibited by technical constraints or legal and privacy considerations \cite{GDPR2016,AIAct,Satyanarayanan2017,Ates:2021ug}. This motivates federated 
	approaches that require only aggregated information about local data to be shared.
	
	Two extreme distributed strategies are unsatisfactory: running independent $k$-means 
	at each device performs poorly for small local datasets, while enforcing 
	identical cluster centroids across all devices via consensus can be suboptimal 
	under statistical heterogeneity.
	
	
	
	We propose a federated $k$-means method that interpolates between these extremes 
	by softly coupling local $k$-means instances over a communication graph. 
	Neighbouring devices are encouraged (but not forced) to agree by penalizing 
	disagreement between their cluster centroids. This yields a formulation of 
	federated clustering as an instance of generalized total variation minimization (GTVMin) \cite{JungFLBook}, and 
	leads to distributed algorithms that operate directly on local cluster centroids.
	 In contrast to methods that train a global set of cluster centroids, we allow 
	 for distinct sets of cluster centroids at well-connected subsets of devices \cite{ClusteredFLTVMinTSP}.

   Generalized total variation minimization (GTVMin) provides a unifying perspective on distributed learning methods such as 
   network Lasso \cite{NetworkLasso}, MOCHA \cite{Smith2017}, and CoCoA \cite{SmithCoCoA}. 
   While recent work applies generalized total variation minimization (GTVMin) to clustered federated learning (CFL) \cite{ClusteredFLTVMinTSP}, 
   we focus on clustering of data points, rather than clustering 
   the devices themselves.
   
   Existing federated $k$-means methods typically compute local cluster centroids and 
   aggregate them into a single global solution \cite{Deng:2025aa}, enforce hard consensus 
   \cite{ForeroJSTSP2011}, or rely on coreset constructions \cite{DistkmeansBalcan2013}. 
   In contrast, we formulate federated $k$-means directly as a generalized total variation minimization (GTVMin) problem over 
   local cluster centroids, enabling flexible, privacy-friendly coordination without central 
   aggregation or coresets.
		 
	 \subsection*{Contribution} 
	
	Our main contributions can be summarized as 
	\begin{itemize} 
		\item (conceptual) we formulate federated $k$-means as an instance of generalized total variation minimization (GTVMin). 
		\item (algorithmic) we obtain a novel federated clustering method by solving the 
		above generalized total variation minimization (GTVMin) instance using a distributed non-linear Jacobi iteration. 
	\end{itemize}

	\subsection*{Notation} 
	We denote Euclidean vectors (matrices) by bold-face lower (upper) letters. 
	The set consisting of the first $m$ natural numbers is denoted $[m] \defeq \{1,\ldots,m\}$. 
	Given a node $\nodeidx\!\in\!\nodes$ of an undirected graph $\graph= \pair{\nodes}{\edges}$, we denote 
	its neighborhood by $\neighbourhood{\nodeidx} \defeq \big\{ \nodeidx' \in \nodes: \edge{\nodeidx}{\nodeidx'} \in \edges \big\}$. The node degree is denoted $\nodedegree{\nodeidx} \defeq \big| \neighbourhood{\nodeidx} \big|$. 
	For a matrix $\mA=\bigg( \va^{(1)},\ldots,\va^{(\samplesize)} \bigg)\in \mathbb{R}^{\nrfeatures \times \samplesize}$, 
	we denote by $\cols{\mA} \defeq \big\{ \va^{(1)},\ldots, \va^{(\samplesize)} \big\} \subseteq \mathbb{R}^{\nrfeatures}$ 
	the set of vectors constituted by the columns of $\mA$.

	\section{Problem Formulation}
	\label{sec_problem_setting} 
	
	We consider a federated learning network (FL network) consisting $\nrnodes$ of devices, which we 
	index by natural numbers $\nodeidx \in \firstnatural{\nrnodes}$. Each device 
	carries its own local dataset $\localdataset{\nodeidx} = \big\{ \featurevec^{(\nodeidx,1)},\ldots, \featurevec^{(\nodeidx,\localsamplesize{\nodeidx})} \big\} \subseteq \mathbb{R}^{\nrfeatures}$. 
	The local datasets constitute the entire pooled dataset 
	\begin{equation} 
		\label{equ_def_pooled_dataset}
		\dataset = \bigcup_{\nodeidx=1}^{\nrnodes} \localdataset{\nodeidx} = \big\{ \featurevec^{(1)},\ldots,\featurevec^{(\samplesize)} \big\}.
	\end{equation} 
	In what follows, we assume that local datasets are pair-wise disjoint and therefore, in turn, 
	$\samplesize = \sum_{\nodeidx=1}^{\nrnodes} \localsamplesize{\nodeidx}$. 

    We want to decompose the pooled dataset \eqref{equ_def_pooled_dataset} 
	into a pre-scribed number $\nrcluster$ of clusters by solving the $k$-means problem 
	\begin{align}
		\min_{\clustercentroid{1},\ldots,\clustercentroid{\nrcluster} \in \mathbb{R}^{\featuredim}} 
			\; \sum_{\sampleidx=1}^{\samplesize} 
				\min_{\clusteridx \in \firstnatural{k}} 
						\normgeneric{ \clustercentroid{\clusteridx}-\featurevec^{(\sampleidx)}}{2}^{2}.
		\label{equ_def_kmeans}
	\end{align} 
	A solution of the $k$-means problem \eqref{equ_def_kmeans} consists of $\nrcluster$ cluster centroids 
	$\widehat{\weights}^{(1)},\ldots,\widehat{\weights}^{(\nrcluster)}$, which we stack into the matrix 
	$$\widehat{\mW} \defeq \big( \widehat{\weights}^{(1)},\ldots,\widehat{\weights}^{(\nrcluster)} \big).$$

	A naive approach would be to collect the entire dataset at a single computational unit (e.g., a server) 
	and then solve \eqref{equ_def_kmeans}. However, this centralized $k$-means approach is 
	infeasible when local datasets contain sensitive information and cannot be shared easily. 
	Instead, we propose a privacy-friendly decentralized federated clustering method. 
	
	Our federated clustering method requires devices to only share aggregated data 
	(e.g., updates of model parameters) across bidirectional communication links. We represent these 
	communication links by the undirected edges $\edges$ of a federated learning network (FL network). Each link $\edge{\nodeidx}{\nodeidx'}\!\in\!\edges$ connects two different devices $\nodeidx, \nodeidx' \in \firstnatural{\nrnodes}$ of the federated learning network (FL network). 
    
	Section \ref{sec_method} presents a de-centralized method for computing approximation of 
	the columns in $\widehat{\mW}$. This method assigns each device $\nodeidx \in \firstnatural{\nrnodes}$ 
	a set of local cluster centroids $\localclustercentroid{1}{\nodeidx},\ldots,\localclustercentroid{\nrcluster}{\nodeidx} \in \mathbb{R}^{\featurelen}$, 
	which we stack into the matrix 
	$$ \mW^{(\nodeidx)} \defeq \big(\localclustercentroid{1}{\nodeidx},\ldots,\localclustercentroid{\nrcluster}{\nodeidx} \big) \in \mathbb{R}^{\featurelen \times k}.$$ 
	This method is iterative: In each iteration, every device updates their local cluster centroids 
	using their local dataset $\localdataset{\nodeidx}$ and the information received via the links $\edges$. 
	
	The goal of our federated $k$-means method is that each device learns the 
    same set of cluster centroids as those obtained by a centralized $k$-means 
    method that solves \eqref{equ_def_kmeans}. To this end, we require 
    \begin{equation}
    	\label{equ_local_consistent_global_kmeans}
    	\cols{\mW^{(\nodeidx)}}\!=\!\cols{\mW^{(\nodeidx')}} \mbox{  for any two devices }\nodeidx, \nodeidx'\!\in\!\firstnatural{\nrnodes}. 
    \end{equation} 
    Note that \eqref{equ_local_consistent_global_kmeans} can hold even 
    if $\mW^{(\nodeidx)} \neq \mW^{(\nodeidx')}$ for some $\nodeidx, \nodeidx' \in \firstnatural{\nrnodes}$. 
    In particular, the validity of \eqref{equ_local_consistent_global_kmeans} does not change 
    if we arbitrarily permute the columns of the matrixs $\mW^{(\nodeidx)}$, for $\nodeidx=1,\ldots,\nrnodes$.
    
	We can measure the extent by which \eqref{equ_local_consistent_global_kmeans} is violated via the generalized total variation (GTV) $\sum_{\edge{\nodeidx}{\nodeidx'} \in \edges} \discrepancy{\nodeidx}{\nodeidx'}$ with 
   	\begin{align} 
   	\label{equ-def_gtv} 
    & \discrepancy{\nodeidx}{\nodeidx'}\!\defeq\!
   	\underbrace{\sum_{\clusteridx \in \firstnatural{\nrcluster}}  \min_{\clusteridx' \in \firstnatural{\nrcluster}} 
   	\normgeneric{\localclustercentroid{\nodeidx}{\clusteridx}\!-\!\localclustercentroid{\nodeidx'}{\clusteridx'}}{2}^2}_{\rm I}
   	 \!+\! \nonumber\\ 
   	 & 	\hspace*{20mm} \underbrace{\sum_{\clusteridx \in \firstnatural{\nrcluster}} \min_{\clusteridx' \in \firstnatural{\nrcluster}} 
   	\normgeneric{\weights^{(\nodeidx',\clusteridx)}\!-\!\weights^{(\nodeidx,\clusteridx')}}{2}^{2}}_{\rm II}.
   	\end{align} 
    Note that, for any link $\edge{\nodeidx}{\nodeidx'}$, we can ensure $\discrepancy{\nodeidx}{\nodeidx'}=0$ 
	only if both components ${\rm I}$ and ${\rm II}$ in \eqref{equ-def_gtv} vanish. 
	Component ${\rm I}$ in \eqref{equ-def_gtv} vanishes if and only if $\cols{\mW^{(\nodeidx)}}  \subseteq \cols{\mW^{(\nodeidx')}}$. 
	Similarly, component ${\rm II}$ in \eqref{equ-def_gtv} vanishes if and only if $\cols{\mW^{(\nodeidx')}}  \subseteq \cols{\mW^{(\nodeidx)}}$. 
	Thus, we have 
    $$  \discrepancy{\nodeidx}{\nodeidx'} = 0 \mbox{ if and only if }     	\cols{\mW^{(\nodeidx)}}\!=\!\cols{\mW^{(\nodeidx')}}.$$
	
	From now on, we assume that the links $\edges$ are such 
	that the resulting undirected graph $\graph = \pair{\firstnatural{\nrnodes}}{\edges}$ 
	is connected. In this case, enforcing a vanishing generalized total variation (GTV) \eqref{equ-def_gtv} 
	implies that each device $\nodeidx \in \firstnatural{\nrnodes}$ 
	carries the same set of local cluster centroids. Thus, by enforcing 
	a vanishing generalized total variation (GTV) \eqref{equ-def_gtv}, the centralized $k$-means 
	problem \eqref{equ_def_kmeans} can be rewritten as 
	\begin{align} 
		\label{equ_def_min_sum_local_kmeans}
	\min_{\mW^{(1)},\ldots,\mW^{(\nrnodes)}}  & \sum_{\nodeidx \in \firstnatural{\nrnodes}} \locallossfunc{\nodeidx}{\mW^{(\nodeidx)}} \mbox{s.t.\ } \sum_{\edge{\nodeidx}{\nodeidx'}\in \edges} 
	 \discrepancy{\nodeidx}{\nodeidx'}=0. 
	\end{align}
	Here, we used the local loss function 
	\begin{align} 
		\locallossfunc{\nodeidx}{\mW} & \defeq \sum_{\sampleidx \in \firstnatural{\localsamplesize{\nodeidx}}} \min_{\clusteridx \in \firstnatural{k}}\normgeneric{\clustercentroid{\clusteridx}\!-\!\featurevec^{(\nodeidx,\sampleidx)}}{2}^{2}  \label{equ_def_local_loss}. 
	\end{align} 
	Note that \eqref{equ_def_local_loss} is a local $k$-means clustering problem 
	for the local dataset $\localdataset{\nodeidx}$. These local $k$-means problems, 
	one for each device $\nodeidx$, are coupled via the constraint 
	$\sum_{\edge{\nodeidx}{\nodeidx'}\in \edges}  \discrepancy{\nodeidx}{\nodeidx'}=0$.

	\section{A Federated $k$-means Method}
	\label{sec_method}
	
	Section \ref{sec_problem_setting} reformulated the basic $k$-means problem 
	\eqref{equ_def_kmeans} as solving \eqref{equ_def_min_sum_local_kmeans} under 
	the constraint of vanishing generalized total variation (GTV) \eqref{equ-def_gtv}. Instead of enforcing a vanishing 
	generalized total variation (GTV) (as in \cite{ForeroJSTSP2011}), we add it as a penalty term to the objective function in \eqref{equ_def_min_sum_local_kmeans}. 
	This results in an instance of generalized total variation minimization (GTVMin) \cite{ClusteredFLTVMinTSP,JungFLBook}, 
	\begin{equation} 
			\min_{\mW^{(1)},\ldots,\mW^{(\nrnodes)}}  \hspace*{-1mm} \sum_{\nodeidx \in \firstnatural{\nrnodes}} \hspace*{-1mm}\locallossfunc{\nodeidx}{\mW^{(\nodeidx)}} 
			\!+\!\regparam \hspace*{-3mm}\sum_{\edge{\nodeidx}{\nodeidx'} \in \edges}\hspace*{-2mm} \discrepancy{\nodeidx}{\nodeidx'}. \label{equ_fedkmeans_gtvmin}
	\end{equation}   
   In general, \eqref{equ_fedkmeans_gtvmin} is a non-convex optimization problem: 
   For the special case $\regparam=0$, \eqref{equ_fedkmeans_gtvmin} decomposes into 
   independent $k$-means problems, one for each device $\nodeidx$. Each 
   of these $k$-means problems is NP hard in general \cite{Mahajan2009Springer,Aloise:2009aa}. 
   
   While solving $k$-means exactly in often infeasible, there is a simple iterative method - 
   referred to as Lloyd's algorithm - for solving $k$-means approximately. This method alternates between assigning data points to nearest cluster centroid and, in turn, re-calculating the cluster centroids given the updated cluster assignment. 
   
   Our main contribution is a distributed variant of the basic Lloyd's algorithm to solve 
   \eqref{equ_fedkmeans_gtvmin}, which yields a federated $k$-means method. 
   The idea is as follows: During the $\iteridx$-th iteration, we update the local cluster centroids 
   $\mW^{(\nodeidx)}$ at some device $\nodeidx$ by minimizing \eqref{equ_fedkmeans_gtvmin} 
   with all other $\mW^{(\nodeidx')}$, $\nodeidx' \in \firstnatural{\nrnodes} \setminus \{ \nodeidx\}$ held fixed. 
   More formally, device $\nodeidx$ replaces its current  $\mW^{(\nodeidx)}$ with a solution of 
   	\begin{align} 
   		\label{equ_def_Jacobi_GTVMin}
   	& \min_{\mW=\big(\clustercentroid{1},\ldots,\clustercentroid{\nrcluster}\big)} \locallossfunc{\nodeidx}{\mW}
   		\!+\!\regparam \hspace*{-3mm}\sum_{\nodeidx' \in \neighbourhood{\nodeidx}} \hspace*{-2mm} \discrepancy{\nodeidx}{\nodeidx'} \nonumber \\ 
   		&\stackrel{\eqref{equ_def_local_loss},\eqref{equ-def_gtv}}{=} \min_{\weights^{(\nodeidx,1)},\ldots,\weights^{(\nodeidx,\nrcluster)}}  
        \frac{1}{\localsamplesize{\nodeidx}} \sum_{\sampleidx \in \firstnatural{\localsamplesize{\nodeidx}}} \min_{\clusteridx \in \firstnatural{\nrcluster}}\normgeneric{\clustercentroid{\clusteridx}- \featurevec^{(\nodeidx,\sampleidx)}}{2}^{2}  \nonumber \\ 
   		& + \alpha \sum_{\nodeidx' \in \neighbourhood{\nodeidx}}
   			\sum_{\clusteridx' \in \firstnatural{\nrcluster}}   \bigg( \underbrace{\min_{\clusteridx \in \firstnatural{\nrcluster}} 
   		\normgeneric{\clustercentroid{\clusteridx}\!-\!\localclustercentroid{\nodeidx'}{\clusteridx'}}{2}^2}_{\rm I} \nonumber \\ 
   		& \hspace*{30mm} \!+\! \underbrace{\min_{\clusteridx \in \firstnatural{\nrcluster}} 
   		\normgeneric{\clustercentroid{\clusteridx'} \!-\!\localclustercentroid{\nodeidx'}{\clusteridx} }{2}^{2}}_{\rm II} \bigg).
   	\end{align}

\begin{algorithm}
	\caption{A Federated $k$-means Method}
	\label{alg:federated_kmeans}
	\begin{algorithmic}[1]
		\Require local datasets $\localdataset{\nodeidx} = \{\featurevec^{(\nodeidx,\sampleidx)}\}_{\sampleidx=1}^{\localsamplesize{\nodeidx}}$, 
		for each device $\nodeidx \in \firstnatural{\nrnodes}$; 
		an undirected graph $\graph=\{\nodes=\firstnatural{\nrnodes},\edges\}$; 
		max.\ nr.\  $K$ of iterations.
		\Ensure local cluster centroids $\mW^{(\iteridx)}$ for all $\nodeidx \in \firstnatural{\nrnodes}$ 
		\State Each device $\nodeidx \in \firstnatural{\nrnodes}$ 
		chooses initial local cluster centroids $\mW^{(\iteridx,1)}$ 
		\For{$\iteridx = 1, 2, \ldots, K$}
			\vspace*{2mm}
		\State Select some device $\nodeidx_{\iteridx}$. \label{alg_fed_kmans_device_select}
		\vspace*{2mm} 
		\State \parbox[t]{0.8\linewidth}{%
			Device $\nodeidx_{\iteridx}$ gathers current local cluster centroids
			$\mW^{(\nodeidx',\iteridx)}$ from its neighbors
			$\nodeidx' \in \neighbourhood{\nodeidx}$.%
		}
			\vspace*{2mm} 
		\State \parbox[t]{0.8\linewidth}{%
        Device $\nodeidx_{\iteridx}$ executes the local steps ({\bf E1})--({\bf E3}) and ({\bf M})
        to obtain $\mW^{(\nodeidx_{\iteridx})}$.%
        }

        \label{alg_fed_kmans_local_update}
		\vspace*{2mm}
		\State Update 
		\begin{align} 
			\mW^{(\nodeidx_{\iteridx},\iteridx+1)}& =\mW^{(\nodeidx_{\iteridx})} \nonumber \\ 
			\mW^{(\nodeidx',\iteridx+1)} &=\mW^{(\nodeidx',\iteridx)} \mbox{ for } \nodeidx' \in \firstnatural{\nrnodes} \setminus \{\nodeidx_{\iteridx} \}. \nonumber 
		\end{align}
		\EndFor
		\State \Return $\mW^{(\nodeidx)}=\mW^{(\nodeidx,K+1)}$ for all $\nodeidx \in \firstnatural{\nrnodes}$.
	\end{algorithmic}
\end{algorithm}

   	The update \eqref{equ_def_Jacobi_GTVMin} bears some resemblance to the $k$-means
   	problem \eqref{equ_def_kmeans}. In particular, for $\alpha=1$ and without the component 
   	 ${\rm II}$, the update \eqref{equ_def_Jacobi_GTVMin} would become exactly a $k$-means 
   	 problem \eqref{equ_def_kmeans} on the augmented local dataset 
   	\begin{equation}
   		\label{equ_augmented_local_dataset}
   		 \bigg\{ \featurevec^{(\nodeidx,1)}, \ldots,\featurevec^{(\nodeidx,\localsamplesize{\nodeidx})}  \bigg\} \bigcup \bigg\{ \vw^{(\nodeidx',\clusteridx')} \bigg\}_{\nodeidx' \in \neighbourhood{\nodeidx},\clusteridx' \in \firstnatural{\nrcluster}}.
   	\end{equation}

We now present our federated $k$-means method Algorithm \ref{alg:federated_kmeans}. This method generates a sequence of local 
cluster centroids at each device. We denote the local cluster centroids at the 
beginning of iteration $\iteridx$ at node $\nodeidx'\!\in\! \firstnatural{\nrnodes}$ by $\mW^{(\nodeidx',\iteridx)}\!=\!\big(\vw^{(\nodeidx',\iteridx,1)},\ldots,\vw^{(\nodeidx',\iteridx,\nrcluster)}\big)$.
During iteration $\iteridx$, Algorithm \ref{alg:federated_kmeans} selects some 
node $\nodeidx = \nodeidx_{\iteridx}$ in step \ref{alg_fed_kmans_device_select} at which an iterative local update scheme is executed to solve the update~\eqref{equ_def_Jacobi_GTVMin} approximately. 

Our local update scheme is a variant of Lloyd's algorithm.
It alternates 
between an cluster centroid-assignment phase and a cluster centroid update step ({\bf M}). 
For given local cluster centroids 
$\mW^{(\nodeidx)}$ and $\mW^{(\nodeidx')} = \mW^{(\nodeidx',\iteridx)}$ for $\nodeidx' \in \neighbourhood{\nodeidx}$, the assignment phase determines the assignment variables at device $\nodeidx$: 
\begin{align}
\hspace*{-2mm}  \hat{\clusteridx}_{\sampleidx} &\!\defeq\!\argmin_{\clusteridx \in \firstnatural{\nrcluster}} 
\norm{\localclustercentroid{\nodeidx}{\clusteridx}\!-\!\featurevec^{(\nodeidx,\sampleidx)}}_{2}^{2} \mbox{ for } \sampleidx\in \firstnatural{\localsamplesize{\nodeidx}}  \nonumber \\ 
\hspace*{-2mm}  \hat{\clusteridx}_{\nodeidx',\clusteridx'} &\!\defeq\!\argmin_{\clusteridx \in \firstnatural{\nrcluster}} 
\norm{\localclustercentroid{\nodeidx}{\clusteridx}\!-\!\vw^{(\nodeidx',\clusteridx')}}_{2}^{2}  \mbox{ for } \nodeidx'\!\in\!\neighbourhood{\nodeidx}, \clusteridx'\!\in\!\firstnatural{\nrcluster} \nonumber \\
\hspace*{-2mm} b_{\nodeidx',\clusteridx'} 
&\!\defeq\!\argmin_{\clusteridx \in \firstnatural{\nrcluster}} 
\norm{\localclustercentroid{\nodeidx}{\clusteridx'}\!-\!\vw^{(\nodeidx',\clusteridx)}}_{2}^{2}  \mbox{ for } \nodeidx'\!\in\!\neighbourhood{\nodeidx}, \clusteridx'\!\in\!\firstnatural{\nrcluster}. \label{equ_def_assignment_aux}
\end{align}
The assignment phase \eqref{equ_def_assignment_aux} consists of three assignment steps,  denoted ({\bf E1}), ({\bf E2}) and ({\bf E3}): 
\begin{itemize} 
\item ({\bf E1}): Each local data point is assigned to the nearest local cluster centroid.
\item ({\bf E2}): For each neighbor cluster centroid $\vw^{(\nodeidx',\clusteridx')}$, for $\clusteridx' \in \firstnatural{\nrcluster}$, the nearest local cluster centroid $\hat{\clusteridx}_{\nodeidx',\clusteridx'} \in \firstnatural{\nrcluster}$
at node $\nodeidx$ is determined.
This step corresponds to solving the minimum ${\rm I}$ in \eqref{equ_def_Jacobi_GTVMin}.
\item ({\bf E3}): For each local cluster centroid $\localclustercentroid{\nodeidx}{\clusteridx}$, 
                         for $\clusteridx \in \firstnatural{\nrcluster}$, we determine the nearest cluster centroid 
                         $b_{\nodeidx',\clusteridx'} \in \firstnatural{\nrcluster}$ at each neighbor $\nodeidx' \in \neighbourhood{\nodeidx}$.
It corresponds to solving the minimum ${\rm II}$ in \eqref{equ_def_Jacobi_GTVMin}.
\end{itemize} 

During the update step ({\bf M}), the local cluster centroids $\mW^{(\nodeidx)} = \big(\widetilde{\weights}^{(\nodeidx,1)},\ldots,\widetilde{\weights}^{(\nodeidx,\nrcluster)}\big) $ are recomputed 
via the weighted averages of locally assigned data points and matched cluster centroids 
at neighbors, scaled by the regularization parameter~$\regparam$:
\begin{equation}
\label{eq:M}
\widetilde{\weights}^{(\nodeidx,\clusteridx)} = \frac{\frac{1}{\localsamplesize{\nodeidx}}\hspace*{-3mm}\sum\limits_{\sampleidx: \hat{\clusteridx}_{\sampleidx}=\clusteridx } 
		\hspace*{-2mm}\featurevec^{(\nodeidx,\sampleidx)}
		\!+\regparam \left[\sum\limits_{(\nodeidx',\clusteridx') \in \mathcal{U}}\hspace*{-4mm}
		\localclustercentroid{\nodeidx'}{\clusteridx'}\!+\hspace*{-3mm}\sum\limits_{\nodeidx' \in \neighbourhood{\nodeidx}} \hspace*{-3mm} \localclustercentroid{\nodeidx'}{b_{\nodeidx',\clusteridx}}\right]}{\max(\frac{1}{\localsamplesize{\nodeidx}}\big| \big\{\sampleidx: \hat{\clusteridx}_{\sampleidx}=\clusteridx \big\} \big|
		\!+\!  \regparam \left[ \big| \mathcal{U} \big|+ \nodedegree{\nodeidx} \right], 1)}.
\end{equation}
In \eqref{eq:M}, $\mathcal{U}=\{(\nodeidx',\clusteridx')\in \neighbourhood{\nodeidx} \times \firstnatural{\nrcluster} : \hat{\clusteridx}_{\nodeidx',\clusteridx'} = \clusteridx \}$ denotes the set of neighbor cluster centroids that are matched to assignment step ({\bf E2}), and $\nodedegree{\nodeidx}$ is the degree of node $\nodeidx$.

Note that the iterative local update scheme in step \ref{alg_fed_kmans_local_update} of Algorithm \ref{alg:federated_kmeans} iteratively alternates between \eqref{equ_def_assignment_aux} and \eqref{eq:M}. It depends 
only on the local cluster centroid $\mW^{(\nodeidx',\iteridx)}$ at neighbors 
$\neighbourhood{\nodeidx}$ of device $\nodeidx$. Breaks when the decrease in \eqref{equ_def_Jacobi_GTVMin} is very small and delivers the last round local cluster centroids $\mW^{({\nodeidx}_{\iteridx})}$.
The device selection in 
step \ref{alg_fed_kmans_device_select} of Algorithm \ref{alg:federated_kmeans} can 
be random or following a deterministic schedule. Any iteration of Algorithm \ref{alg:federated_kmeans} can never increase the objective function in \eqref{equ_fedkmeans_gtvmin}, i.e., Algorithm \ref{alg:federated_kmeans} always
converges in terms of \eqref{equ_fedkmeans_gtvmin}.

When $\regparam=1$ and the second matching step ({\bf E3}) is omitted (corresponding to 
removing component ${\rm II}$ in \eqref{equ_def_Jacobi_GTVMin}), this iterative update scheme becomes standard $k$-means \cite[Ch. 8]{MLBasics} applied to the augmented local dataset 
\eqref{equ_augmented_local_dataset}.

\section{Numerical Experiments}
\label{eu_sec:experiments}
In this section, we present numerical experiments to evaluate the effectiveness of Algorithm \ref{alg:federated_kmeans}. 
Section \ref{eu_experiment_setup} details the experimental setup, while section \ref{eu_local_sample}, section \ref{eu_graph_connectivity} and section \ref{eu_discussion} presents the results and their discussion. 

\subsection{Experimental Setup} 
\label{eu_experiment_setup}
\newcommand{\GCD}{\mathrm{GCD}} 
\newcommand{\CV}{\mathrm{CV}}   
\newcommand{\LD}{\mathrm{LD}}   
\newcommand{\datasetiso}{\mathcal{D}_{\mathrm{iso}}}
\newcommand{\datasetvar}{\mathcal{D}_{\mathrm{var}}}
\newcommand{\datasetaniso}{\mathcal{D}_{\mathrm{aniso}}}

We evaluate Algorithm \ref{alg:federated_kmeans} and all baselines on three synthetic datasets from scikit-learn, representing different cluster geometries:
\emph{Isotropic}~$\datasetiso$ with cluster variance $\sigma=1.0$. \emph{Heterogeneous-Variance} $\datasetvar$ uses cluster-specific variances $\sigma_{\clusteridx} \in \{1.0, 2.5, 0.5\}$ for $\clusteridx \in \firstnatural{\nrcluster}$.
\emph{Anisotropic}~$\datasetaniso$ obtained via a fixed rotation–scaling transform $\mathbf{A}=\begin{bmatrix}0.6 & -0.6 \\ -0.4 & 0.8\end{bmatrix}$.
All datasets contain $\nrcluster=3$ clusters in $\realcoorspace{2}$ with minimum centroid separation $5$ and are uniformly distributed across $\nrnodes=10$ nodes. 
The communication topology is modeled as an Erdős–Rényi graph with edge probability $p$.

We benchmark against distributed $k$-means in \cite{ForeroJSTSP2011}, and local $k$-means (no communication). We use two metrics to evaluate the clustering performance: 
\begin{align}
\hspace*{-2mm}
\GCD
&\!\defeq\!
\frac{1}{2 \nrnodes \nrcluster}
\sum_{\nodeidx \in \nodes}
\discrepancy{\nodeidx}{\mathrm{cen}}
\nonumber
\\
\hspace*{-2mm}
\CV
&\!\defeq\!
\frac{1}{2 \nrcluster \, n_{\mathrm{eff}}}
\sum_{|\neighbourhood{\nodeidx}|>0}
\frac{1}{|\neighbourhood{\nodeidx}|}
\sum_{\nodeidx' \in \neighbourhood{\nodeidx}}
\discrepancy{\nodeidx}{\nodeidx'}.
\label{equ_def_gcd_cv}
\end{align}

Note that \emph{Global Centroid Deviation (GCD)} quantifies the deviation between the local solutions 
$\widehat{\mW}^{(\nodeidx)} = \{\localclustercentroid{\nodeidx}{\clusteridx}\}_{\clusteridx=1}^{\nrcluster}$
and the centralized $k$-means solution $\widehat{\mW}^{\mathrm{cen}} = \{\localclustercentroid{\mathrm{cen}}{\clusteridx'}\}_{\clusteridx'=1}^{\nrcluster}$ 
on the pooled dataset, and \emph{Consensus Variation (CV)} measures cross-device clustering variation over node pair $(\nodeidx,\nodeidx')$ in graph $\graph$. $\discrepancy{\nodeidx}{\mathrm{cen}}$ and $\discrepancy{\nodeidx}{\nodeidx'}$ 
follow the definition in \eqref{equ-def_gtv}, and $\nrnodes_{\text{eff}}$ denotes the number of nodes with at least one neighbor.
All methods are initialized with $k$-means++, run for $T=200$ iterations, and 
evaluated over $10$ independent runs, with error bars indicating the standard 
error. Source code of our experiments is available at  \href{https://version.aalto.fi/gitlab/junga1/flclustering/-/blob/cc4e7dea7640835d62aa29fc26b94f1bf19f34e4/FedKmeans/NumExpXu/FedKMeans_GTVmin.ipynb}{Gitlab}. The rows in Figures \ref{fig:central_error_alpha}--\ref{fig:node_error_p} correspond to different datasets; the vertical axis uses logarithmic scale.

\begin{figure}[h!]
	\centering
	\makebox[\linewidth][c]{%
		\scalebox{0.9}{\input{fig1.tex}}
	}
	\caption{Global centroid deviation (GCD) of Algorithm~\ref{alg:federated_kmeans}, distributed $k$-means, and local $k$-means.}
	\label{fig:central_error_alpha}
\end{figure}

\subsection{Performance comparison under different local sample size} 
\label{eu_local_sample}

We study the effect of local sample size by varying $\localsamplesize{\nodeidx}\in \{50, ..., 800\}$ in increments of $50$ with fixed connectivity $p=0.7$. Distributed $k$-means is run with $\eta=2$, while Algorithm~\ref{alg:federated_kmeans} is evaluated for $\alpha\in\{0,0.5,1\}$.
Figure \ref{fig:central_error_alpha} shows that increasing $\alpha$ progressively reduces GCD, with the proposed method approaching the centralized solution as local sample size grows. For $\alpha=1$, Algorithm \ref{alg:federated_kmeans} outperforms distributed $k$-means on the isotropic 
and heterogeneous-variance datasets with lower GCD 
on anisotropic data, where distributed $k$-means 
may fall behind local $k$-means.

Figure \ref{fig:node_error_alpha} indicates that Algorithm~\ref{alg:federated_kmeans} achieves 
decreasing cross-node disagreement as local sample size increases, whereas distributed $k$-means 
exhibits the opposite trend. 

\begin{figure}[h!]
	\centering
	\makebox[\linewidth][c]{%
		\scalebox{0.9}{\input{fig2.tex}}
	}
	\caption{Consensus Variation (CV) of Algorithm \ref{alg:federated_kmeans} and distributed $k$-means.}
	\label{fig:node_error_alpha}
\end{figure}
\vspace*{-2mm}
\subsection{Impact of Connectivity} 
\label{eu_graph_connectivity}

We evaluate the impact of graph connectivity by varying the edge probability $p\in\{0.4,0.7,1\}$. 
Figure \ref{fig:central_error_baselines} shows that Algorithm \ref{alg:federated_kmeans} 
achieves decreasing GCD as connectivity increases, reaching the best performance under 
full connectivity across all datasets. In contrast, distributed $k$-means exhibits unstable 
behavior, particularly on anisotropic data and under sparse graphs, where it falls behind 
local $k$-means. Figure \ref{fig:node_error_p} reports the corresponding CV. 
Algorithm \ref{alg:federated_kmeans} achieves stronger consensus as local sample 
size increases, largely independent of graph connectivity, whereas distributed $k$-means 
is highly sensitive to sparse communication and shows degraded consensus. 

\begin{figure}[h!]
	\centering
	\makebox[\linewidth][c]{%
		\scalebox{0.9}{\input{fig3.tex}}
	}
	\caption{Global Centroid Deviation (GCD) of Algorithm \ref{alg:federated_kmeans} and baselines for varying connectivity.}
	\label{fig:central_error_baselines}
\end{figure}

\begin{figure}[htbp]
	\centering
	\makebox[\linewidth][c]{%
		\scalebox{0.9}{\input{fig4.tex}}
	}
	\caption{Consensus Variation (CV) of Algorithm \ref{alg:federated_kmeans} and distributed $k$-means for varying connectivity.}
	\label{fig:node_error_p}
\end{figure}

\subsection{Discussion}
\label{eu_discussion}

Both Algorithm \ref{alg:federated_kmeans} and distributed $k$-means address decentralized clustering by coupling local $k$-means objectives through graph-based interactions. While both methods rely on iterative updates over an undirected communication graph, they differ fundamentally in their optimization formulations.

Distributed $k$-means enforces exact consensus among connected nodes via ``hard'' 
equality constraints and is solved using a primal-dual method. In contrast, the proposed 
method operates directly on the cluster centroids and promotes agreement 
via penalizing generalized total variation (GTV), without requiring exact consensus. This ``soft'' coupling allows 
our method to better adapt to the connectivity of a federated learning network (FL network). The results of our 
numerical experiments verify the usefulness of generalized total variation minimization (GTVMin)-based 
method for decentralized federated clustering. 

  \vspace*{-2mm}
\section{Acknowledgements}
Ekkehard Schnoor and Mahsa Asadi reviewed early drafts of 
the manuscript. 
\vspace*{-2mm}

\bibliographystyle{IEEEtran}

\end{document}

%% file: ml_macros.tex
 \usepackage{mleftright}
 
\newcommand{\discrepancy}[2]{d^{(#1,#2)}}  
\newcommand{\firstnatural}[1]{[#1]}    

\newcommand{\defeq}{:=}

\newcommand{\vx}[0]{{\bf x}}

\newcommand{\mA}[0]{{\bf A}}

\newcommand{\mW}[0]{{\bf W}}

\newcommand{\vw}[0]{{\bf w}}

\newcommand{\va}[0]{{\bf a}}

\newcommand{\cols}[1]{{\rm cols}\left(#1 \right)}  

\newcommand{\neighbourhood}[1]{\mathcal{N}^{(#1)}}

\newcommand{\norm}[1]{\Vert  {#1} \Vert}
\newcommand{\normgeneric}[2]{\mleft\lVert #1 \mright\rVert_{#2}}

\newcommand{\bmx}[0]{\begin{bmatrix}}
\newcommand{\emx}[0]{\end{bmatrix}}

\newcommand{\featuredim}{d}
\newcommand{\nrfeatures}{\featuredim}   

\newcommand{\featurelen}{\featuredim}     

\newcommand{\samplesize}{m}
\newcommand{\sampleidx}{r}

\newcommand{\clusteridx}{c} 
\newcommand{\nrcluster}{k}   
  
\newcommand{\localclustercentroid}[2]{{\vw}^{(#1,#2)}}   
\newcommand{\clustercentroid}[1]{{\vw}^{(#1)}} 
\newcommand{\clustercentroiditer}[2]{{\vw}^{(#1,#2)}} 

\newcommand{\featurevec}{\vx}

\newcommand{\dataset}{\mathcal{D}}

\newcommand{\realcoorspace}[1]{\mathbb{R}^{\text{#1}}}

\newcommand{\regparam}{\alpha}

\DeclareMathOperator*{\argmin}{arg\,min}

\newcommand{\iteridx}{t}

\newcommand{\weights}{\vw}

\newcommand{\locallossfunc}[2]{L_{#1}\left(#2 \right)}

\newcommand{\localdataset}[1]{\mathcal{D}^{(#1)}}

\newcommand{\edges}{\mathcal{E}}

\newcommand{\graph}{\mathcal{G}}
\newcommand{\nodes}{\mathcal{V}}

\newcommand{\nodedegree}[1]{d^{(#1)}}

\newcommand{\nodeidx}{i}
\newcommand{\nrnodes}{n}

\newcommand{\edge}[2]{\{#1,#2\}}

\newcommand{\localsamplesize}[1]{m_{#1}}

\newcommand{\pair}[2]{\left( #1,#2 \right)}


%% file: fig1.tex
\pgfplotsset{
	algfed/.style  ={mark=*, thick, dashed, mark size=1.2pt,
		error bars/.cd, y explicit, y dir=both,
		error bar style={thin, solid},
		error mark=- , error mark options={line width=1.2pt, mark size=2pt, rotate=0 }
	},
	alglocal/.style={mark=*, thick, dashdotted, mark size=1.2pt,
		error bars/.cd, y explicit, y dir=both,
		error bar style={thin, solid},
		error mark=- , error mark options={line width=1.2pt, mark size=2pt}
	},
    algdist/.style={mark=triangle*, thick, dotted, mark size=1.2pt,
        error bars/.cd, y explicit, y dir=both,
        error bar style={thin, solid},
        error mark=- , error mark options={line width=1.2pt, mark size=2pt, rotate=0 }}
}

\begin{tikzpicture}[remember picture]
	\begin{groupplot}[
		group style={
			group size=3 by 3, 
			horizontal sep=0.5cm,
			vertical sep=0.2cm,    
			y descriptions at=edge left,
			ylabels at=edge left,
			yticklabels at=edge left,
			x descriptions at=edge bottom,
			xlabels at=edge bottom,
			xticklabels at=edge bottom,
			group name=myplots
		},
		height=4cm,
		width=0.22\textwidth,
		ymin=1e-7, 
		ymax=10,
		enlarge y limits=upper,
		ylabel style={align=center, yshift=0cm}, 
		ymode=log,
		ticklabel style={font=\footnotesize},
		cycle list name=mycolors,
		legend style={draw=none, at={(-0.5,-0.4)}, anchor=north, nodes={scale=0.55}, legend columns=-1, font=\Large},
		legend image post style={scale=0.4},
		axis x line*=bottom,
		axis y line*=left,
		xmin=0, xmax=800,
		xtick={0,200,400,600,800},
        xticklabel style = {rotate=-45, anchor= west}
		]
		
		\nextgroupplot[title={\small $\alpha=0$},
        ylabel={\normalsize $\GCD, \datasetiso$},
        ymin=1e-32, 
		ymax=10]
		\addplot+[algfed
		] table[
		col sep=comma,
		x index=0, y index=1, y error index=4
		] {tikz_tables/results_blobs_alpha_0_0.csv};
		
		\addplot+[alglocal
		] table[
		col sep=comma,
		x index=0, y index=2, y error index=5
		] {tikz_tables/results_blobs_alpha_0_0.csv};

        \addplot+[algdist
		] table[
		col sep=comma,
		x index=0, y index=3, y error index=6
		] {tikz_tables/results_blobs_alpha_0_0.csv};
		
		\nextgroupplot[title={\small $\alpha=0.5$},
        ymin=1e-32, 
		ymax=10
		]
		\addplot+[algfed
		] table[
		col sep=comma,
		x index=0, y index=1, y error index=3
		] {tikz_tables/results_blobs_alpha_0_5.csv};
		\addplot+[alglocal
		] table[
		col sep=comma,
		x index=0, y index=2, y error index=5
		] {tikz_tables/results_blobs_alpha_0_5.csv};

        \addplot+[algdist
		] table[
		col sep=comma,
		x index=0, y index=3, y error index=6
		] {tikz_tables/results_blobs_alpha_0_5.csv};
		
		\nextgroupplot[title={\small $\alpha=1$},
        ymin=1e-32, 
		ymax=10
		]
		\addplot+[algfed
		] table[
		col sep=comma,
		x index=0, y index=1, y error index=4
		] {tikz_tables/results_blobs_alpha_1_0.csv};
		
		\addplot+[alglocal
		] table[
		col sep=comma,
		x index=0, y index=2, y error index=5
		] {tikz_tables/results_blobs_alpha_1_0.csv};
        \addplot+[algdist
		] table[
		col sep=comma,
		x index=0, y index=3, y error index=6
		] {tikz_tables/results_blobs_alpha_1_0.csv};
		
		\nextgroupplot[ylabel={\normalsize $\GCD, \datasetaniso$}]
		\addplot+[algfed
		] table[
		col sep=comma,
		x index=0, y index=1, y error index=4
		] {tikz_tables/results_aniso_alpha_0_0.csv};
		
		\addplot+[alglocal
		] table[
		col sep=comma,
		x index=0, y index=2, y error index=5
		] {tikz_tables/results_aniso_alpha_0_0.csv};

        \addplot+[algdist
		] table[
		col sep=comma,
		x index=0, y index=3, y error index=6
		] {tikz_tables/results_aniso_alpha_0_0.csv};
		
		\nextgroupplot[
		]
		\addplot+[algfed
		] table[
		col sep=comma,
		x index=0, y index=1, y error index=4
		] {tikz_tables/results_aniso_alpha_0_5.csv};
		\addplot+[alglocal
		] table[
		col sep=comma,
		x index=0, y index=2, y error index=5
		] {tikz_tables/results_aniso_alpha_0_5.csv};
        \addplot+[algdist
		] table[
		col sep=comma,
		x index=0, y index=3, y error index=6
		] {tikz_tables/results_aniso_alpha_0_5.csv};
		
		\nextgroupplot[
		]
		\addplot+[algfed
		] table[
		col sep=comma,
		x index=0, y index=1, y error index=4
		] {tikz_tables/results_aniso_alpha_1_0.csv};
		
		\addplot+[alglocal
		] table[
		col sep=comma,
		x index=0, y index=2, y error index=5
		] {tikz_tables/results_aniso_alpha_1_0.csv};
		\addplot+[algdist
		] table[
		col sep=comma,
		x index=0, y index=3, y error index=6
		] {tikz_tables/results_aniso_alpha_1_0.csv};
		
		\nextgroupplot[ 
        ylabel={\normalsize $\GCD, \datasetvar$}]
		\addplot+[algfed
		] table[
		col sep=comma,
		x index=0, y index=1, y error index=4
		] {tikz_tables/results_varied_alpha_0_0.csv};
		
		\addplot+[alglocal
		] table[
		col sep=comma,
		x index=0, y index=2, y error index=5
		] {tikz_tables/results_varied_alpha_0_0.csv};
        
        \addplot+[algdist
		] table[
		col sep=comma,
		x index=0, y index=3, y error index=6
		] {tikz_tables/results_varied_alpha_0_0.csv};
		
		\nextgroupplot[xlabel={$\localsamplesize{\nodeidx}$},
        xlabel style={font=\small}]
		\addplot+[algfed
		] table[
		col sep=comma,
		x index=0, y index=1, y error index=3
		] {tikz_tables/results_varied_alpha_0_5.csv};
		\addplot+[alglocal
		] table[
		col sep=comma,
		x index=0, y index=2, y error index=4
		] {tikz_tables/results_varied_alpha_0_5.csv};
        
        \addplot+[algdist
		] table[
		col sep=comma,
		x index=0, y index=3, y error index=6
		] {tikz_tables/results_varied_alpha_0_5.csv};
		
		\nextgroupplot[
		legend entries={
        $Algorithm\, \ref{alg:federated_kmeans}$, $ Local\, k\textit{-}means$, $Distributed\, k\textit{-}means$ } 
		]
		\addplot+[algfed
		] table[
		col sep=comma,
		x index=0, y index=1, y error index=4
		] {tikz_tables/results_varied_alpha_1_0.csv};
		
		\addplot+[alglocal
		] table[
		col sep=comma,
		x index=0, y index=2, y error index=5
		] {tikz_tables/results_varied_alpha_1_0.csv};

        \addplot+[algdist
		] table[
		col sep=comma,
		x index=0, y index=3, y error index=6
		] {tikz_tables/results_varied_alpha_1_0.csv};
		
	\end{groupplot}
\end{tikzpicture}

%% file: fig2.tex
\pgfplotsset{
	algfed/.style  ={mark=*, thick, dashed, mark size=1.2pt,
		error bars/.cd, y explicit, y dir=both,
		error bar style={thin, solid},
		error mark=- , error mark options={line width=1.2pt, mark size=2pt, rotate=0 }
	},
    algdist/.style={mark=triangle*, thick, dotted, mark size=1.2pt, color=green!50!black, 
        error bars/.cd, y explicit, y dir=both,
        error bar style={thin, solid}, 
        error mark=- , error mark options={line width=1.2pt, mark size=2pt, rotate=0 }}
}

\begin{tikzpicture}[remember picture]
	\begin{groupplot}[
		group style={
			group size=3 by 3, 
			horizontal sep=0.5cm,
			vertical sep=0.2cm,    
			y descriptions at=edge left,
			ylabels at=edge left,
			yticklabels at=edge left,
			x descriptions at=edge bottom,
			xlabels at=edge bottom,
			xticklabels at=edge bottom,
			group name=myplots
		},
		height=4cm,
		width=0.22\textwidth,
		ymin=1e-6, 
		ymax=10,
		enlarge y limits=upper,
		ylabel style={align=center, yshift=0cm}, 
		ymode=log,
		ticklabel style={font=\footnotesize},
		cycle list name=mycolors,
		legend style={draw=none, at={(-0.5,-0.1)}, anchor=north, nodes={scale=0.55}, legend columns=-1, font=\Large},
		legend image post style={scale=0.4},
		axis x line*=bottom,
		axis y line*=left,
		xmin=0, xmax=800,
		xtick={0,200,400,600,800},
        xticklabel style = {rotate=-45, anchor=west}
		]
		
		\nextgroupplot[title={\small $\alpha = 0$},
        ylabel={\normalsize $\CV, \datasetiso$},
        ymin=1e-32]
		\addplot+[algfed
		] table[
		col sep=comma,
		x index=0, y index=7, y error index=10
		] {tikz_tables/results_blobs_alpha_0_0.csv};
        \addplot+[algdist
		] table[
		col sep=comma,
		x index=0, y index=9, y error index=12
		] {tikz_tables/results_blobs_alpha_0_0.csv};
		
		\nextgroupplot[title={\small $\alpha = 0.5$},
        ymin=1e-32
		]
		\addplot+[algfed
		] table[
		col sep=comma,
		x index=0, y index=7, y error index=10
		] {tikz_tables/results_blobs_alpha_0_5.csv};
        \addplot+[algdist
		] table[
		col sep=comma,
		x index=0, y index=9, y error index=12
		] {tikz_tables/results_blobs_alpha_0_5.csv};
		
		\nextgroupplot[title={\small $\alpha = 1$},
        ymin=1e-32
		]
		\addplot+[algfed
		] table[
		col sep=comma,
		x index=0, y index=7, y error index=10
		] {tikz_tables/results_blobs_alpha_1_0.csv};
        \addplot+[algdist
		] table[
		col sep=comma,
		x index=0, y index=9, y error index=12
		] {tikz_tables/results_blobs_alpha_1_0.csv};
		
		\nextgroupplot[ylabel={\normalsize $\CV, \datasetaniso$},
        ymin=1e-16]
		\addplot+[algfed
		] table[
		col sep=comma,
		x index=0, y index=7, y error index=10
		] {tikz_tables/results_aniso_alpha_0_0.csv};
        \addplot+[algdist
		] table[
		col sep=comma,
		x index=0, y index=9, y error index=12
		] {tikz_tables/results_aniso_alpha_0_0.csv};
		
		\nextgroupplot[
        ymin=1e-16
		]
		\addplot+[algfed
		] table[
		col sep=comma,
		x index=0, y index=7, y error index=10
		] {tikz_tables/results_aniso_alpha_0_5.csv};
        \addplot+[algdist
		] table[
		col sep=comma,
		x index=0, y index=9, y error index=12
		] {tikz_tables/results_aniso_alpha_0_5.csv};
		
		\nextgroupplot[
        ymin=1e-16
		]
		\addplot+[algfed
		] table[
		col sep=comma,
		x index=0, y index=7, y error index=10
		] {tikz_tables/results_aniso_alpha_1_0.csv};
        \addplot+[algdist
		] table[
		col sep=comma,
		x index=0, y index=9, y error index=12
		] {tikz_tables/results_aniso_alpha_1_0.csv};
		
		\nextgroupplot[
        ylabel={\normalsize $\CV, \datasetvar$},
        ymin=1e-32]
		\addplot+[algfed
		] table[
		col sep=comma,
		x index=0, y index=7, y error index=10
		] {tikz_tables/results_varied_alpha_0_0.csv};
        \addplot+[algdist
		] table[
		col sep=comma,
		x index=0, y index=9, y error index=12
		] {tikz_tables/results_varied_alpha_0_0.csv};
		
		\nextgroupplot[ xlabel={\small $\localsamplesize{\nodeidx}$},
        ymin=1e-32]
		\addplot+[algfed
		] table[
		col sep=comma,
		x index=0, y index=7, y error index=10
		] {tikz_tables/results_varied_alpha_0_5.csv};
        \addplot+[algdist
		] table[
		col sep=comma,
		x index=0, y index=9, y error index=12
		] {tikz_tables/results_varied_alpha_0_5.csv};
        
		\nextgroupplot[
		legend entries={$Algorithm\, \ref{alg:federated_kmeans}$,$ Distributed \, k\textit{-}means$}, 
		legend style={draw=none, at={(-0.5,-0.4)}, anchor=north, legend columns=-1, font=\Large},
        ymin=1e-32
		]
		\addplot+[algfed
		] table[
		col sep=comma,
		x index=0, y index=7, y error index=10
		] {tikz_tables/results_varied_alpha_1_0.csv};
        \addplot+[algdist
		] table[
		col sep=comma,
		x index=0, y index=9, y error index=12
		] {tikz_tables/results_varied_alpha_1_0.csv};
		
	\end{groupplot}
	\tikzset{SubCaption/.style={
			text width=2in, yshift=0mm, align=center, anchor=north, outer sep=0.5cm,
	}}
	\tikzset{SubCaptionBottom/.style={
			text width=3in, yshift=-2mm, align=center, anchor=north, outer sep=1cm,
	}}
\end{tikzpicture}

%% file: fig3.tex
\pgfplotsset{
	algfed/.style  ={mark=*, thick, dashed, mark size=1.2pt,
		error bars/.cd, y explicit, y dir=both,
		error bar style={thin, solid},
		error mark=- , error mark options={line width=1.2pt, mark size=2pt, rotate=0 }
	},
	alglocal/.style={mark=*, thick, dashdotted, mark size=1.2pt,
		error bars/.cd, y explicit, y dir=both,
		error bar style={thin, solid},
		error mark=- , error mark options={line width=1.2pt, mark size=2pt}
	},
    algdist/.style={mark=triangle*, thick, dotted, mark size=1.2pt,
        error bars/.cd, y explicit, y dir=both,
        error bar style={thin, solid},
        error mark=- , error mark options={line width=1.2pt, mark size=2pt, rotate=0 }}
}

\begin{tikzpicture}[remember picture]
	\begin{groupplot}[
		group style={
			group size=3 by 3,  
			horizontal sep=0.5cm,
			vertical sep=0.2cm,    
			y descriptions at=edge left,
			ylabels at=edge left,
			yticklabels at=edge left,
			x descriptions at=edge bottom,
			xlabels at=edge bottom,
			xticklabels at=edge bottom,
			group name=myplots
		},
		height=4cm,
		width=0.22\textwidth,
		ymin=1e-7, 
		ymax=10,
		enlarge y limits=upper,
		ylabel style={align=center, yshift=0cm}, 
		ymode=log,
		ticklabel style={font=\footnotesize},
		cycle list name=mycolors,
		legend style={draw=none, at={(-0.5,-0.1)}, anchor=north, nodes={scale=0.55}, legend columns=-1, font=\Large},
		legend image post style={scale=0.4},
		axis x line*=bottom,
		axis y line*=left,
		xmin=0, xmax=800,
		xtick={0,200,400,600,800},
        xticklabel style = {rotate=-45, anchor=west}
		]
		
		\nextgroupplot[title={\small $p=0.4$},
        ylabel={\normalsize $\GCD, \datasetiso$},
        ymin=1e-32]
		\addplot+[algfed
		] table[
		col sep=comma,
		x index=0, y index=1, y error index=2
		] {tikz_tables/errcentral_blobs_p0_4.csv};
		
		\addplot+[alglocal
		] table[
		col sep=comma,
		x index=0, y index=3, y error index=4
		] {tikz_tables/errcentral_blobs_p0_4.csv};

        \addplot+[algdist
		] table[
		col sep=comma,
		x index=0, y index=5, y error index=6
		] {tikz_tables/errcentral_blobs_p0_4.csv};
		
		\nextgroupplot[title={\small $p=0.7$},
        ymin=1e-32
		]
		\addplot+[algfed
		] table[
		col sep=comma,
		x index=0, y index=1, y error index=2
		] {tikz_tables/errcentral_blobs_p0_7.csv};
		\addplot+[alglocal
		] table[
		col sep=comma,
		x index=0, y index=3, y error index=4
		] {tikz_tables/errcentral_blobs_p0_7.csv};
        \addplot+[algdist
		] table[
		col sep=comma,
		x index=0, y index=5, y error index=6
		] {tikz_tables/errcentral_blobs_p0_7.csv};
		
		\nextgroupplot[title={\small $p=1$},
        ymin=1e-32
		]
		\addplot+[algfed
		] table[
		col sep=comma,
		x index=0, y index=1, y error index=2
		] {tikz_tables/errcentral_blobs_p1_0.csv};
		
		\addplot+[alglocal
		] table[
		col sep=comma,
		x index=0, y index=3, y error index=4
		] {tikz_tables/errcentral_blobs_p1_0.csv};

        \addplot+[algdist
		] table[
		col sep=comma,
		x index=0, y index=5, y error index=6
		] {tikz_tables/errcentral_blobs_p1_0.csv};
		
		\nextgroupplot[ylabel={\normalsize $\GCD, \datasetaniso$}]
		\addplot+[algfed
		] table[
		col sep=comma,
		x index=0, y index=1, y error index=2
		] {tikz_tables/errcentral_aniso_p0_4.csv};
		
		\addplot+[alglocal
		] table[
		col sep=comma,
		x index=0, y index=3, y error index=4
		] {tikz_tables/errcentral_aniso_p0_4.csv};
        \addplot+[algdist
		] table[
		col sep=comma,
		x index=0, y index=5, y error index=6
		] {tikz_tables/errcentral_aniso_p0_4.csv};
		
		\nextgroupplot[ymin=1e-9
		]
		\addplot+[algfed
		] table[
		col sep=comma,
		x index=0, y index=1, y error index=2
		] {tikz_tables/errcentral_aniso_p0_7.csv};
		\addplot+[alglocal
		] table[
		col sep=comma,
		x index=0, y index=3, y error index=4
		] {tikz_tables/errcentral_aniso_p0_7.csv};
        \addplot+[algdist
		] table[
		col sep=comma,
		x index=0, y index=5, y error index=6
		] {tikz_tables/errcentral_aniso_p0_7.csv};
		
		\nextgroupplot[ymin=1e-9
		]
		\addplot+[algfed
		] table[
		col sep=comma,
		x index=0, y index=1, y error index=2
		] {tikz_tables/errcentral_aniso_p1_0.csv};
		
		\addplot+[alglocal
		] table[
		col sep=comma,
		x index=0, y index=3, y error index=4
		] {tikz_tables/errcentral_aniso_p1_0.csv};
        \addplot+[algdist
		] table[
		col sep=comma,
		x index=0, y index=5, y error index=6
		] {tikz_tables/errcentral_aniso_p1_0.csv};

		\nextgroupplot[
        ylabel={\normalsize $\GCD, \datasetvar$}, ymax=1e3, ymin=1e-9]
		\addplot+[algfed
		] table[
		col sep=comma,
		x index=0, y index=1, y error index=2
		] {tikz_tables/errcentral_varied_p0_4.csv};
		
		\addplot+[alglocal
		] table[
		col sep=comma,
		x index=0, y index=3, y error index=4
		] {tikz_tables/errcentral_varied_p0_4.csv};
        \addplot+[algdist
		] table[
		col sep=comma,
		x index=0, y index=5, y error index=6
		] {tikz_tables/errcentral_varied_p0_4.csv};
		
		\nextgroupplot[xlabel={\small $\localsamplesize{\nodeidx}$}, ymax=1e3]
		\addplot+[algfed
		] table[
		col sep=comma,
		x index=0, y index=1, y error index=2
		] {tikz_tables/errcentral_varied_p0_7.csv};
		\addplot+[alglocal
		] table[
		col sep=comma,
		x index=0, y index=3, y error index=4
		] {tikz_tables/errcentral_varied_p0_7.csv};
        \addplot+[algdist
		] table[
		col sep=comma,
		x index=0, y index=5, y error index=6
		] {tikz_tables/errcentral_varied_p0_7.csv};
		
		\nextgroupplot[
		legend entries={$Algorithm\, \ref{alg:federated_kmeans}$, $Local\, k\textit{-}means$, $Distributed\, k\textit{-}means$}, 
		legend style={draw=none, at={(-0.5,-0.4)}, anchor=north, legend columns=-1, font=\Large}, ymax=1e3
		]
		\addplot+[algfed
		] table[
		col sep=comma,
		x index=0, y index=1, y error index=2
		] {tikz_tables/errcentral_varied_p1_0.csv};
		
		\addplot+[alglocal
		] table[
		col sep=comma,
		x index=0, y index=3, y error index=4
		] {tikz_tables/errcentral_varied_p1_0.csv};
        \addplot+[algdist
		] table[
		col sep=comma,
		x index=0, y index=5, y error index=6
		] {tikz_tables/errcentral_varied_p1_0.csv};
		
	\end{groupplot}
	\tikzset{SubCaption/.style={
			text width=2in, yshift=0mm, align=center, anchor=north, outer sep=0.5cm,
	}}
	\tikzset{SubCaptionBottom/.style={
			text width=3in, yshift=-2mm, align=center, anchor=north, outer sep=1cm,
	}}
\end{tikzpicture}

%% file: fig4.tex
\pgfplotsset{
	algfed/.style  ={mark=*, thick, dashed, mark size=1.2pt,
		error bars/.cd, y explicit, y dir=both,
		error bar style={thin, solid},
		error mark=- , error mark options={line width=1.2pt, mark size=2pt, rotate=0 }
	},
    algdist/.style={mark=triangle*, thick, dotted, mark size=1.2pt, color=green!50!black, 
        error bars/.cd, y explicit, y dir=both,
        error bar style={thin, solid},
        error mark=- , error mark options={line width=1.2pt, mark size=2pt, rotate=0 }}
}

\begin{tikzpicture}[remember picture]
	\begin{groupplot}[
		group style={
			group size=3 by 3,  
			horizontal sep=0.5cm,
			vertical sep=0.2cm,    
			y descriptions at=edge left,
			ylabels at=edge left,
			yticklabels at=edge left,
			x descriptions at=edge bottom,
			xlabels at=edge bottom,
			xticklabels at=edge bottom,
			group name=myplots
		},
		height=4cm,
		width=0.22\textwidth,
		ymin=1e-20, 
		ymax=10,
		enlarge y limits=upper,
		ylabel style={align=center, yshift=0cm}, 
		ymode=log,
		ticklabel style={font=\footnotesize},
		cycle list name=mycolors,
		legend style={draw=none, at={(-0.5,-0.1)}, anchor=north, nodes={scale=0.55}, legend columns=-1, font=\Large},
		legend image post style={scale=0.4},
		axis x line*=bottom,
		axis y line*=left,
		xmin=0, xmax=800,
		xtick={0,200,400,600,800},
        xticklabel style = {rotate=-45, anchor=west}
		]
		
		\nextgroupplot[title={\small $p=0.4$},
        ylabel={\normalsize $\CV, \datasetiso$},
        ymin=1e-32]
		\addplot+[algfed
		] table[
		col sep=comma,
		x index=0, y index=1, y error index=2
		] {tikz_tables/errneib_blobs_p0_4.csv};
		
		\addplot+[algdist
		] table[
		col sep=comma,
		x index=0, y index=5, y error index=6
		] {tikz_tables/errneib_blobs_p0_4.csv};
		
		\nextgroupplot[title={\small $p=0.7$},
        ymin=1e-32]
		\addplot+[algfed
		] table[
		col sep=comma,
		x index=0, y index=1, y error index=2
		] {tikz_tables/errneib_blobs_p0_7.csv};
		\addplot+[algdist
		] table[
		col sep=comma,
		x index=0, y index=5, y error index=6
		] {tikz_tables/errneib_blobs_p0_7.csv};
		
		\nextgroupplot[title={\small $p=1$},
        ymin=1e-32
		]
		\addplot+[algfed
		] table[
		col sep=comma,
		x index=0, y index=1, y error index=2
		] {tikz_tables/errneib_blobs_p1_0.csv};
		
		\addplot+[algdist
		] table[
		col sep=comma,
		x index=0, y index=5, y error index=6
		] {tikz_tables/errneib_blobs_p1_0.csv};

		\nextgroupplot[ylabel={\normalsize $\CV, \datasetaniso$},
        ymin=1e-32]
		\addplot+[algfed
		] table[
		col sep=comma,
		x index=0, y index=1, y error index=2
		] {tikz_tables/errneib_aniso_p0_4.csv};
		
		\addplot+[algdist
		] table[
		col sep=comma,
		x index=0, y index=5, y error index=6
		] {tikz_tables/errneib_aniso_p0_4.csv};
		
		\nextgroupplot[ymin=1e-32]
		\addplot+[algfed
		] table[
		col sep=comma,
		x index=0, y index=1, y error index=2
		] {tikz_tables/errneib_aniso_p0_7.csv};
		\addplot+[algdist
		] table[
		col sep=comma,
		x index=0, y index=5, y error index=6
		] {tikz_tables/errneib_aniso_p0_7.csv};
		
		\nextgroupplot[
        ymin=1e-32
		]
		\addplot+[algfed
		] table[
		col sep=comma,
		x index=0, y index=1, y error index=2
		] {tikz_tables/errneib_aniso_p1_0.csv};
		
		\addplot+[algdist
		] table[
		col sep=comma,
		x index=0, y index=5, y error index=6
		] {tikz_tables/errneib_aniso_p1_0.csv};

		\nextgroupplot[ylabel={\normalsize $\CV, \datasetvar$},
        ymin=1e-32]
		\addplot+[algfed
		] table[
		col sep=comma,
		x index=0, y index=1, y error index=2
		] {tikz_tables/errneib_varied_p0_4.csv};
		
		\addplot+[algdist
		] table[
		col sep=comma,
		x index=0, y index=5, y error index=6
		] {tikz_tables/errneib_varied_p0_4.csv};
		
		\nextgroupplot[xlabel={$\localsamplesize{\nodeidx}$},
        xlabel style={font=\small},
        ymin=1e-32]
		\addplot+[algfed
		] table[
		col sep=comma,
		x index=0, y index=1, y error index=2
		] {tikz_tables/errneib_varied_p0_7.csv};
		\addplot+[algdist
		] table[
		col sep=comma,
		x index=0, y index=5, y error index=6
		] {tikz_tables/errneib_varied_p0_7.csv};
		
		\nextgroupplot[
		legend entries={$Algorithm\, \ref{alg:federated_kmeans}$, $Distributed\, k\textit{-}means$}, 
		legend style={draw=none, at={(-0.5,-0.4)}, anchor=north, legend columns=-1, font=\Large},
        ymin=1e-32
		]
		\addplot+[algfed
		] table[
		col sep=comma,
		x index=0, y index=1, y error index=2
		] {tikz_tables/errneib_varied_p1_0.csv};
		
		\addplot+[algdist
		] table[
		col sep=comma,
		x index=0, y index=5, y error index=6
		] {tikz_tables/errneib_varied_p1_0.csv};
	\end{groupplot}
	\tikzset{SubCaption/.style={
			text width=2in, yshift=0mm, align=center, anchor=north, outer sep=0.5cm,
	}}
	\tikzset{SubCaptionBottom/.style={
			text width=3in, yshift=-2mm, align=center, anchor=north, outer sep=1cm,
	}}
\end{tikzpicture}

%% file: FedKMeansGTVMin_Arxiv.bbl
\begin{thebibliography}{10}
	\providecommand{\url}[1]{#1}
	\csname url@samestyle\endcsname
	\providecommand{\newblock}{\relax}
	\providecommand{\bibinfo}[2]{#2}
	\providecommand{\BIBentrySTDinterwordspacing}{\spaceskip=0pt\relax}
	\providecommand{\BIBentryALTinterwordstretchfactor}{4}
	\providecommand{\BIBentryALTinterwordspacing}{\spaceskip=\fontdimen2\font plus
		\BIBentryALTinterwordstretchfactor\fontdimen3\font minus
		\fontdimen4\font\relax}
	\providecommand{\BIBforeignlanguage}[2]{{%
			\expandafter\ifx\csname l@#1\endcsname\relax
			\typeout{** WARNING: IEEEtran.bst: No hyphenation pattern has been}%
			\typeout{** loaded for the language `#1'. Using the pattern for}%
			\typeout{** the default language instead.}%
			\else
			\language=\csname l@#1\endcsname
			\fi
			#2}}
	\providecommand{\BIBdecl}{\relax}
	\BIBdecl
	
	\bibitem{MLBasics}
	A.~Jung, \emph{Machine Learning: The Basics}.\hskip 1em plus 0.5em minus
	0.4em\relax Singapore, Singapore: Springer Nature, 2022.
	
	\bibitem{hastie01statisticallearning}
	T.~Hastie, R.~Tibshirani, and J.~Friedman, \emph{The Elements of Statistical
		Learning: Data Mining, Inference, and Prediction}, 2nd~ed.\hskip 1em plus
	0.5em minus 0.4em\relax New York, NY, USA: Springer Science+Business Media,
	2009.
	
	\bibitem{AaltoDictofML}
	A.~Jung, K.~Olioumtsevits, E.~Schnoor, T.~F. Ryyn{\"a}nen, J.~Gronier, and
	S.~Rastelli, ``The aalto dictionary of machine learning,'' 2025.
	
	\bibitem{BishopBook}
	C.~M. Bishop, \emph{Pattern Recognition and Machine Learning}.\hskip 1em plus
	0.5em minus 0.4em\relax New York, NY, USA: Springer Science+Business Media,
	2006.
	
	\bibitem{Aloise:2009aa}
	\BIBentryALTinterwordspacing
	D.~Aloise, A.~Deshpande, P.~Hansen, and P.~Popat, ``Np-hardness of euclidean
	sum-of-squares clustering,'' \emph{Machine Learning}, vol.~75, no.~2, pp.
	245--248, 2009. [Online]. Available:
	\url{https://doi.org/10.1007/s10994-009-5103-0}
	\BIBentrySTDinterwordspacing
	
	\bibitem{Lloyd1982}
	S.~Lloyd, ``Least squares quantization in {{PCM}},'' \emph{IEEE Trans. Inf.
		Theory}, vol.~28, no.~2, pp. 129--137, Mar. 1982.
	
	\bibitem{Gray1980}
	R.~Gray, J.~Kieffer, and Y.~Linde, ``Locally optimal block quantizer design,''
	\emph{Information and Control}, vol.~45, pp. 178--198, 1980.
	
	\bibitem{GDPR2016}
	\BIBentryALTinterwordspacing
	{European Parliament} and {Council of the European Union}, ``Regulation ({EU})
	2016/679 of the {European Parliament and of the Council of 27 April 2016} on
	the protection of natural persons with regard to the processing of personal
	data and on the free movement of such data, and repealing {Directive
		95/46/EC} ({General Data Protection Regulation}) ({Text with EEA
		relevance}),'' {Official Journal of the European Union, L~119/1, May 4},
	2016, {Accessed: July, 2025}. [Online]. Available:
	\url{https://eur-lex.europa.eu/eli/reg/2016/679/oj}
	\BIBentrySTDinterwordspacing
	
	\bibitem{AIAct}
	\BIBentryALTinterwordspacing
	{European Commission}, ``Proposal for a regulation of the {E}uropean
	{P}arliament and of the {C}ouncil laying down harmonised rules on artificial
	intelligence ({A}rtificial {I}ntelligence {A}ct) and amending certain {U}nion
	legislative acts,'' {COM/2021/206 final, Apr. 21}, 2021, {Accessed: December
		16, 2025}. [Online]. Available:
	\url{https://eur-lex.europa.eu/legal-content/EN/TXT/?uri=celex:52021PC0206}
	\BIBentrySTDinterwordspacing
	
	\bibitem{Satyanarayanan2017}
	\BIBentryALTinterwordspacing
	M.~Satyanarayanan, ``The emergence of edge computing,'' \emph{Computer},
	vol.~50, no.~1, pp. 30--39, Jan. 2017. [Online]. Available:
	\url{https://doi.org/10.1109/MC.2017.9}
	\BIBentrySTDinterwordspacing
	
	\bibitem{Ates:2021ug}
	\BIBentryALTinterwordspacing
	H.~C. Ates, A.~K. Yetisen, F.~G{\"u}der, and C.~Dincer, ``Wearable devices for
	the detection of covid-19,'' \emph{Nature Electronics}, vol.~4, no.~1, pp.
	13--14, 2021. [Online]. Available:
	\url{https://doi.org/10.1038/s41928-020-00533-1}
	\BIBentrySTDinterwordspacing
	
	\bibitem{JungFLBook}
	A.~Jung, \emph{Federated Learning: From Theory to Practice}.\hskip 1em plus
	0.5em minus 0.4em\relax Springer Nature, Jan. 2026.
	
	\bibitem{ClusteredFLTVMinTSP}
	Y.~SarcheshmehPour, Y.~Tian, L.~Zhang, and A.~Jung, ``Clustered federated
	learning via generalized total variation minimization,'' \emph{IEEE Trans.
		Signal Process.}, vol.~71, pp. 4240--4256, 2023, doi:
	10.1109/TSP.2023.3322848.
	
	\bibitem{NetworkLasso}
	D.~Hallac, J.~Leskovec, and S.~Boyd, ``Network lasso: Clustering and
	optimization in large graphs,'' in \emph{Proc. SIGKDD}, 2015, pp. 387--396.
	
	\bibitem{Smith2017}
	\BIBentryALTinterwordspacing
	V.~Smith, C.-K. Chiang, M.~Sanjabi, and A.~Talwalkar, ``Federated
	{M}ulti-{T}ask {L}earning,'' in \emph{Advances in Neural Information
		Processing Systems}, vol.~30, 2017. [Online]. Available:
	\url{https://proceedings.neurips.cc/paper/2017/file/6211080fa89981f66b1a0c9d55c61d0f-Paper.pdf}
	\BIBentrySTDinterwordspacing
	
	\bibitem{SmithCoCoA}
	\BIBentryALTinterwordspacing
	V.~Smith, S.~Forte, C.~Ma, M.~Tak{{\'a}}{\v{c}}, M.~I. Jordan, and M.~Jaggi,
	``{CoCoA}: A general framework for communication-efficient distributed
	optimization,'' \emph{Journal of Machine Learning Research}, vol.~18, no.
	230, pp. 1--49, 2018. [Online]. Available:
	\url{http://jmlr.org/papers/v18/16-512.html}
	\BIBentrySTDinterwordspacing
	
	\bibitem{Deng:2025aa}
	Z.~Deng, Y.~Wang, and M.~M. Alobaedy, ``\BIBforeignlanguage{eng}{Federated
		k-means based on clusters backbone.}'' \emph{\BIBforeignlanguage{eng}{PLoS
			One}}, vol.~20, no.~6, p. e0326145, 2025.
	
	\bibitem{ForeroJSTSP2011}
	P.~A. Forero, A.~Cano, and G.~B. Giannakis, ``Distributed clustering using
	wireless sensor networks,'' \emph{IEEE Journal of Selected Topics in Signal
		Processing}, vol.~5, no.~4, pp. 707--724, 2011.
	
	\bibitem{DistkmeansBalcan2013}
	\BIBentryALTinterwordspacing
	M.~F. Balcan, S.~Ehrlich, and Y.~Liang, ``Distributed k-means and k-median
	clustering on general topologies,'' in \emph{Adv. Neural Inf. Process.
		Syst.}, C.~J. Burges, L.~Bottou, M.~Welling, Z.~Ghahramani, and K.~Q.
	Weinberger, Eds., vol.~26, 2013, pp. 1995--2003. [Online]. Available:
	\url{https://papers.nips.cc/paper_files/paper/2013/hash/7f975a56c761db6506eca0b37ce6ec87-Abstract.html}
	\BIBentrySTDinterwordspacing
	
	\bibitem{Mahajan2009Springer}
	M.~Mahajan, P.~Nimbhorkar, and K.~Varadarajan, ``The planar k-means problem is
	{{NP}}-hard,'' in \emph{WALCOM: Algorithms and Computation}, S.~Das and
	R.~Uehara, Eds.\hskip 1em plus 0.5em minus 0.4em\relax Berlin, Heidelberg,
	Germany: Springer-Verlag, 2009, pp. 274--285.
	
\end{thebibliography}
